%% file: ms.tex
\def\year{2021}\relax

\documentclass[letterpaper]{article}

\usepackage{aaai21} 

\nocopyright
\usepackage{times} 
\usepackage{helvet}
\usepackage{placeins}
\usepackage{courier} 
\usepackage[hyphens]{url} 
\usepackage{graphicx} 
\usepackage{natbib} 
\usepackage{caption} 
\title{WRSE - a non-parametric weighted-resolution ensemble for\\ predicting individual survival distributions in the ICU}
\author{
   Jonathan Heitz  \textsuperscript{\rm 1},
   Joanna Ficek \textsuperscript{\rm 1,4,5},
   Martin Faltys \textsuperscript{\rm 1,2}, 
   Tobias M. Merz \textsuperscript{\rm 2,3}, \\
   Gunnar Rätsch \textsuperscript{\rm 1,5},
   Matthias Hüser \textsuperscript{\rm 1,5} \\
}
\affiliations {
    \textsuperscript{\rm 1} Department of Computer Science, ETH Zürich, Switzerland \\
    \textsuperscript{\rm 2} Department of Intensive Care Medicine, University Hospital Bern, Switzerland \\
    \textsuperscript{\rm 3} Cardiovascular Intensive Care Unit, Auckland City Hospital, New Zealand \\
    \textsuperscript{\rm 4} PhD Program Molecular \& Translational Biomedicine, Life Science Zurich Graduate School, Switzerland \\
    \textsuperscript{\rm 5} Swiss Institute of Bioinformatics, Lausanne, Switzerland \\
}

\usepackage{amsmath}
\usepackage{placeins}
\usepackage{amsfonts}

\begin{document}

\maketitle

\input{abstract}

\input{introduction}
\input{related_work}
\input{real_time_survival_dist}
\input{evaluation_metrics}
\input{prob_time_to_death}
\input{experiments}
\input{discussion}

\input{acknowledgements}

\bibliography{refs}

\input{appendix}

\end{document}

%% file: abstract.tex
\begin{abstract}
Dynamic assessment of mortality risk in the intensive care unit (ICU) can be used to stratify patients, inform about treatment effectiveness or serve as part of an early-warning system. Static risk scoring systems, such as APACHE or SAPS, have recently been supplemented with data-driven approaches that track the dynamic mortality risk over time. Recent works have focused on enhancing the information delivered to clinicians even further by producing full survival distributions instead of point predictions or fixed horizon risks. In this work, we propose a non-parametric ensemble model, \emph{Weighted Resolution Survival Ensemble} (\emph{WRSE}), tailored to estimate such
dynamic individual survival distributions. Inspired by the simplicity and robustness of ensemble methods, the proposed approach combines a set of binary classifiers spaced according to a decay function reflecting the relevance of short-term mortality predictions. Models and baselines are evaluated under weighted calibration and discrimination metrics for individual survival distributions which closely reflect the utility of a model in ICU practice. We show competitive results with state-of-the-art probabilistic models, while greatly reducing training time by factors of 2-9x.
\end{abstract}

%% file: introduction.tex
\section{Introduction}

Mortality prediction in the intensive care unit (ICU) has historically used scores, such as APACHE~II \citep{knaus1985apacheII} or SAPS \citep{legall1983saps}, which group patients into broad risk categories using data from the beginning of their stay \citep{keegan2011severityreview}. More recently,
\emph{dynamic} mortality prediction models were proposed
which solve a binary classification problem with a fixed horizon, e.g., the next 24 hours \citep{maslove2018severity}, which limits their flexibility. Other approaches only provide point predictions of individual survival times and these have been shown to be too imprecise for clinical practice \citep{henderson2001accuracyofpointpredictionsinsurvivalanalysis, henderson2005individualsurvivaltimeprediction}. For effective decision making in an ICU setting, a more expressive and flexible risk estimate is desirable, for instance predicting the full distribution over the remaining time-to-death \citep{avati2018countdownRegression,haider2020evaluation}. 
In this manner, individual patients with high or increasing mortality risk could be identified to direct the physician's attention. Decreasing mortality risk, on the other hand, would provide reassurance to clinicians that the chosen treatment is effective. In addition, resource management in ICUs has become a pressing issue to allocate the limited treatment capacity \citep{Murthy2020-zt}. Thus, a continuously updated mortality risk for current and potential ICU patients could support continuous triaging.

To address this need, several parametric \citep{avati2018countdownRegression} as well non-parametric \citep{lee2018deephit, yu2011mtlr} models have been proposed. In this work, inspired by the success of ensemble and stacking models in the context of survival analysis \citep{pirracchio2015superlearner, wey2015ensemble, lee2019tempquilting}, we propose a non-parametric approach that combines predictions of a set of classifiers using a decaying weight function controlling
the temporal spacing and hence resolution in future
time horizons. We call our approach \emph{Weighted Resolution Survival Ensemble} (\emph{WRSE}). By choosing particular
decay functions, our ensemble gives more importance to short-term predictions most relevant in various ICU settings. We evaluate our models and baselines under weighted evaluation metrics that capture temporal calibration and discrimination.

%% file: related_work.tex
\section{Related work}

The most commonly used approach for predicting survival curves, the Kaplan-Meier estimator \citep{kaplan1958}, provides class-specific information on a population level, but cannot be used for producing individual survival distributions over time. To address this issue, several approaches, both parametric and non-parametric, have been proposed \citep{ishwaran2008rsf, yu2011mtlr, avati2018countdownRegression}. 
Furthermore, recent approaches \citep{lee2018deephit, kvamme2019continuous} have leveraged the power of neural networks to analyze vast amounts of data and account for possible non-linear relationships to produce a full survival distribution per patient. 

Several works have employed the strategy of combining different survival models to benefit from their strengths and obtain superior results to each model considered separately. \cite{pirracchio2015superlearner} build a \emph{Super Learner}, an ensemble of binary classifiers, such as APACHE II or SAPS, to predict in-hospital death. This method, however, does not allow for estimation of survival curves. Other approaches are based on \emph{stacking}, i.e. combining either predictor matrices or predictions from survival models. In the first case, the predictor matrices of patients in the risk set at a given failure time point are concatenated together with the risk set indicator and a binary classifier is applied to the stacked matrix, yielding the conditional probability of experiencing the event at each considered time point \citep{zhong2019survival}. When combined with regression models, the approach can be used to estimate survival curves. In the second case, stacking survival models corresponds to providing a weighted combination of survival function estimates, with time-independent \cite{wey2015ensemble} or time-dependent weighting \citep[\emph{temporal quilting}]{lee2019tempquilting}. Such approaches are prone to overfitting because additional meta-parameters, more of them for time-dependent weighting, are introduced. Moreover, they suffer from the practical drawback of the training time being determined by 
the slowest base model.

On the topic of evaluation metrics for survival prediction, \cite{cook2006performanceassessment} argued that
the key properties of a survival model are discrimination and calibration. Discrimination reflects 
the correct ordering of patients by the estimated probability of death. To take into account 
the time component, \cite{antolini2005time} introduced a \emph{time-dependent discrimination index}, 
an adaptation of Harrell's C index \citep{harrell1982evaluating} that operates directly on predicted
survival functions, instead of relying on point estimates. Calibration captures how well a model's predictions reflect the true frequency of the events. \cite{haider2020evaluation} and \cite{avati2018countdownRegression} generalized the notion of calibration to the full survival distribution. For both calibration and discrimination,
recently proposed metrics ignore the relative importance of different future time horizons in an ICU. To bridge this gap, we propose weighted metrics that support selection of the best model for this particular application.

%% file: real_time_survival_dist.tex
\section{Dynamic individual survival distributions}

We consider a temporal dataset $\{ \{(\mathbf{x}_t^i,c^i,y_t^i) \}_{t=1}^{k_i} \}_{i=1}^N$ of $N$ patients with (partially known) times-to-death $T_t^i$, and $k_i$ the number of observed (hourly) time-points of patient $i$. Let $\mathbf{x}^i_t \in \mathbb{R}^d$ represent the multi-dimensional feature vector, incorporating all available information for a patient $i$ at a certain point in time $t$. Let $c^i \in \{0, 1\}$ be an indicator variable where $c=1$ denotes censoring and $c=0$ means that the patient died in the ICU (non-censored). $y_t^i$ is the observed time-to-death (if $c=0$, then $T_t^i=y_t^i$) or time to discharge (if $c=1$, then $T_t^i>y_t^i$). The latter may take place both in case of improvement (discharge to another hospital unit) and worsening (transfer to palliative care) of the patient's status, and the only information, with respect to the event, is patient's survival (lack of event) until the end of the ICU stay. Therefore, the discharge event represents censoring.
The task is to predict the distribution of the time-to-death $T_t^i$ for all time-steps $t$ during the ICU stay. We use $\hat{F_t^i}(\tau)$ to denote the predicted cumulative distribution function (CDF) for all future times $\tau>0$, as viewed from $t$. $\hat{S}_t^i(\tau) = 1 - \hat{F_t^i}(\tau)$ describes the predicted survival function.

%% file: evaluation_metrics.tex
\section{Weighted discrimination/calibration metrics suitable
         for the intensive care unit}
         
\label{sec:evaluation_metrics}
\begin{figure}[h!]
  \centering
  \includegraphics[width=0.9\columnwidth]{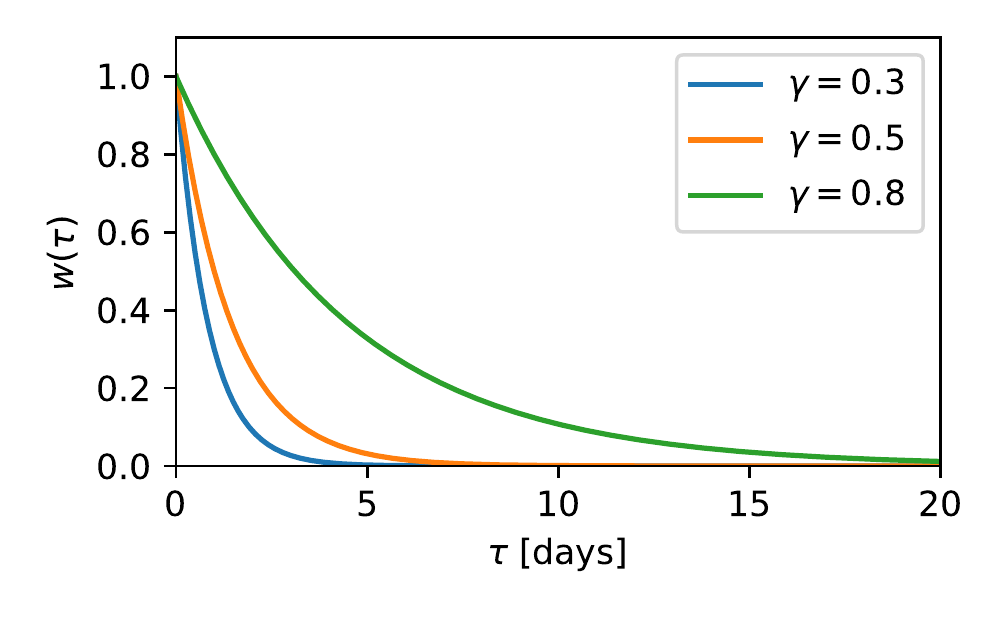}
  \caption{The exponentially decaying weighting function $w(t) = \gamma^t$ for $\gamma \in \{0.3, 0.5, 0.8\}$ .}
  \label{fig:weighting_function}
\end{figure}

We consider two metrics to measure performance of models in the ICU setting, each of which can be calculated for every future point in time $\tau$: calibration, and discrimination. To adapt these metrics to better reflect clinical usefulness of the models, we weight short-term predictions more strongly, using an exponentially decaying weighting function $w(\tau) = \gamma^\tau$ with rate $\gamma \in (0,1)$ and $\tau$ measured in days. The function $w(\tau) = \gamma^\tau$ is plotted for parameters $\gamma=0.3$, $\gamma=0.5$ and $\gamma=0.8$ in Fig.~\ref{fig:weighting_function}. The three settings were chosen as examples of 
strong weighting of the short-term future (next 2 days), as well as more moderate decay with some relevance for the medium-term and long-term future. Depending on the desired application and its time horizon, other choices of $\gamma$ may be appropriate. 

We evaluate calibration of the estimated probabilities $Pr[T<\tau]$ for every time $\tau$ using a calibration curve, plotting mean predicted values in predefined bins against the respective fraction of positives. This produces a piece-wise linear function $c_\tau(q)$ with $c_\tau(0)=0$ and $c_\tau(1)=1$. The diagonal $c^*(q)=q$ represents perfect calibration. We report the absolute area around the diagonal as a metric of calibration for every $\tau$: $\int_0^1 |c_\tau(q) - c^*(q)| \; dq$. Patients who do not die are not considered for $\tau$ after their time of discharge. The \emph{weighted absolute area around the diagonal} \emph{Cal}$^w$ is then given as a weighted mean: 
\begin{equation*}
    \text{\emph{Cal}}^w = \frac{\sum_{\tau \in T} w(\tau) \cdot \int_0^1 |c_\tau(q) - c^*(q)| \; dq}{\sum_{\tau \in T} w(\tau)} \ .
\end{equation*}

To evaluate discrimination, we adapt the time-dependent discrimination index $C^{td}$ \citep{antolini2005time}. Let $\mathcal{T}$ be the set of all distinct times of  death ($\mathcal{T} = \{y^k | c^k = 0\} $). For each $\tau \in \mathcal{T}$, we consider  the set of pairs of patients $\mathcal{P}_\tau = \{(i,j) | y^i = \tau \land c^i = 0 \land y^j \geq \tau\}$. Here, patient $i$ has died at time $\tau$ and patient $j$ is still at risk, i.e. has not died or been discharged until time $\tau$. A pair is concordant if $\hat{S}^i(y^i)\leq\hat{S}^j(y^i)$ and $C^{td}$ 
averages the fraction of concordant pairs over time. The \emph{weighted time-dependent discrimination index} $C^{td,w}$ is then: 
\begin{small}
\begin{equation*}
    C^{td,w} = \frac{\sum_{\tau \in \mathcal{T}}{|\{(i,j) \in \mathcal{P}_\tau | \hat{S}^i(\tau) < \hat{S}^j(\tau)\}| \cdot w(\tau)}} {\sum_{\tau \in \mathcal{T}}{|P_\tau| \cdot w(\tau)}} \ .
\end{equation*}
\end{small}

%% file: prob_time_to_death.tex
\section{WRSE and baseline models}
\label{sec:prob_time_to_death}

\subsection{WRSE (Weighted Resolution Survival Ensemble)}
\label{subsec:WRSE}

\begin{figure}[h!]
  \centering
  \includegraphics[width=0.9\columnwidth]{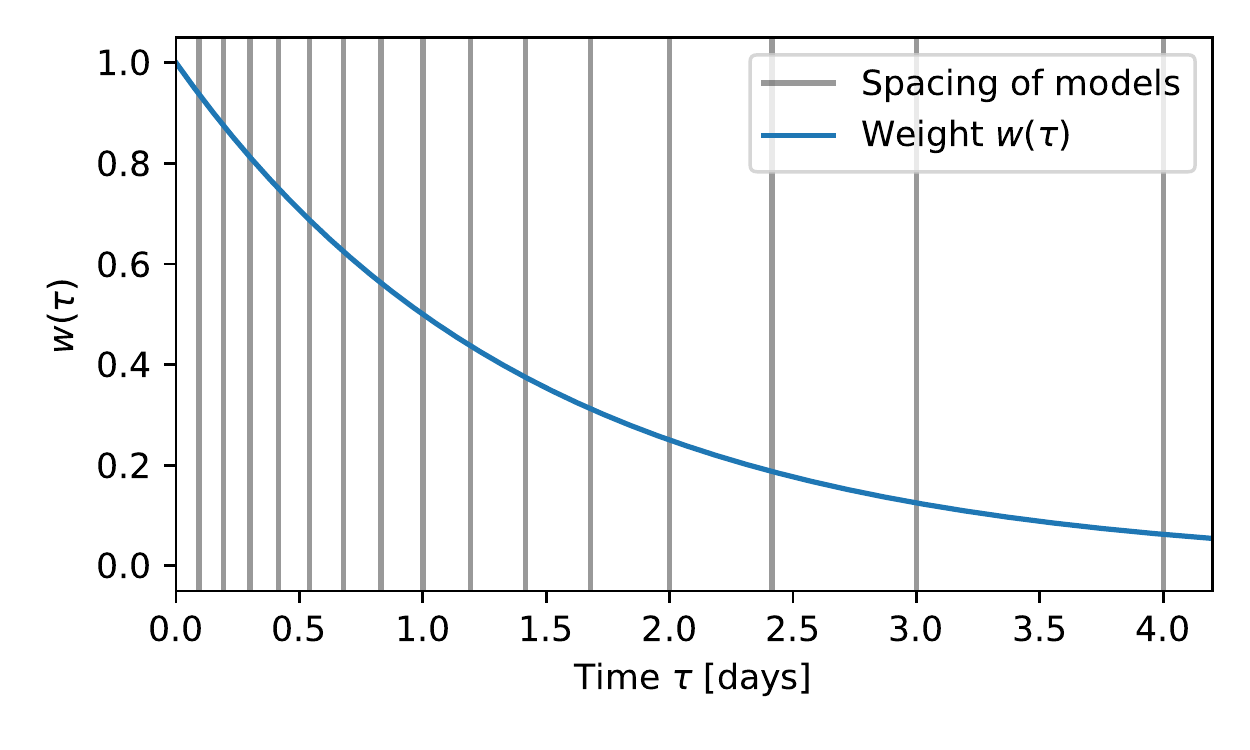}
  \caption{The exponentially decaying weighting function $w(t) = \gamma^\tau$ for $\gamma=0.5$ as well as the spacing of 15 base models in our ensemble method according to this weighting.}
  \label{fig:model_spacing}
\end{figure}

We propose a non-parametric ensemble estimator, which we call
\emph{WRSE}. It consists of binary classification base models $m_1, \dots , m_{K}$, where $m_k(\mathbf{x})$ for $k \in \{1, \dots, K\}$ predicts the probability of dying within the next $h_k$ hours ($Pr[T < h_k  \; | \; \mathbf{x}]$), given patient information $\mathbf{x}$ and an increasing sequence $h = (h_1, \dots, h_K)$. We use $K=15$ and define $h_k = w^{-1}(1-\frac{k}{K+1})$, where $w^{-1}(\cdot) = \log(\cdot) / \log(\gamma)$ is the inverse of $w(\cdot)$ defined in Section~\ref{sec:evaluation_metrics}, thereby putting more emphasis on the short-term future. This spacing of base models is depicted in Fig.~\ref{fig:model_spacing}. We combine the base model predictions by interpreting them as non-parametric estimates of the cumulative distribution function (CDF) for $0 \leq T \leq h_K$. We do not predict the shape of the CDF for $T > h_K$, as such long term horizons are rarely present in the ICU. Due to the independence of the base models, the sequence $m_1(\mathbf{x}), \dots, m_K(\mathbf{x})$ is not guaranteed to be monotonically increasing for a given patient, a necessary condition for a CDF. We solve this using the isotonic regression framework \citep{barlow1972isotonic}, finding an optimal fit vector $m^*_1, \dots, m^*_K$ subject to monotonicity constraints. More precisely, isotonic regression solves the following constrained optimization problem:
\begin{align*}
    &\min \sum_{k=1}^K (m^*_k - m_k(\mathbf{x}))^2 \\
    &\text{subject to } m^*_1 \leq m^*_2 \leq \dots \leq m^*_K \ .
\end{align*}

The ensemble framework is general, and any binary classification model could be used. We employ \emph{LightGBM} because of its demonstrated high performance for ICU data \citep{hyland2019circulatoryfailure} and
interpretability using the TreeSHAP algorithm \cite{lundberg2018consistent}. Each
\emph{LightGBM} base model has at most 64 leaves in each tree, at most 1000 trees, and a learning rate of 0.01.

\subsection{Parametric survival prediction baselines}\label{sec:parametric}
As a baseline, we evaluate two parametric models (\emph{Log-normal} and \emph{Exponential}) to predict $\hat{F}_t$, given $\mathbf{x}_t$, both based on the Survival-CRPS loss function \citep{avati2018countdownRegression}. The framework allows to incorporate the information from censored observations into the model, with the loss function given by
\begin{small}
\begin{align*} 
	\mathcal{S}_{CRPS}\left(\hat{F}_t, (y,c)\right) = &\int_0^y \hat{F}_t(\tau)^2 \;d\tau \\ + (1-c) &\int_y^\infty \left( 1 - \hat{F}_t(\tau) \right) ^2 \;d\tau \ .
	%\label{eq:survival_crps}
\end{align*} 
\end{small}
 The log-normal distribution is commonly chosen in survival analysis due to its flexibility \citep{avati2018countdownRegression,royston2001lognormal,yang2017lognormal}.
For this distribution, a closed-form representation of $\mathcal{S}_{CRPS}$ does not exist and thus, we use a trapezoidal approximation suitable for backpropagation \citep{avati2018countdownRegression}. In addition, we evaluate the exponential distribution, motivated by a preliminary analysis suggesting good fit of this distribution to our ICU data-set.

The cumulative density function is given by $\hat{F_t}^{\lambda}(\tau) = 1-\exp(-\lambda \cdot \tau)$ for a parameter $\lambda$. $\mathcal{S}_{CRPS}$ then has a closed-form representation, which is derived as follows:

\begin{small}
\begin{align*}
	& \mathcal{S}_{CRPS}\left(\hat{F_t}^\lambda, (y,c)\right) \\
	= &\int_0^y \hat{F_t}^\lambda(\tau)^2 \;d\tau + (1-c) \int_y^\infty \left( 1 - \hat{F_t}^\lambda(\tau) \right) ^2 \;d\tau \\ 
	= &\int_0^y (1-e^{-\lambda \tau})^2 \;d\tau + (1-c) \int_y^\infty e^{-2 \lambda \tau} \;d\tau \\
	= &\frac{4 e^{-\lambda y} - c e^{-2 \lambda y } - 3}{2 \lambda} + y \ .
\end{align*}
\end{small}

For these parametric models, we analyzed two feature extractor choices. 
A multi-layer perceptron (MLP) with two hidden layers as well as temporal convolutional network (TCN) blocks, which have been shown
to outperform RNN-based models in sequence modeling tasks \citep{bai2018empirical, franceschi2019unsupervised}. All models were trained using an Adam optimizer \citep{kingma2014adam} and 
early stopping,  terminating training when the validation loss does not improve for 10 epochs. Two hidden layers of 50 neurons each, moderate weight regularization of 0.01, and a learning rate of 1e-4 was used. The output layer directly predicts one (exponential) or two (log-normal) distribution parameters. Results on the performance of the two
feature extractor choices in these models are shown in Appendix Table \ref{tab:results_feature_extractor}.

\subsection{Non-parametric survival prediction baselines}

To avoid making strong assumptions about the underlying distribution, more flexible non-parametric models are commonly employed to model survival distributions \citep{lee2018deephit, kvamme2019continuous, alaa2017nonparametric}.

Recently, several approaches based on neural networks have been proposed. \cite{lee2018deephit} introduced \emph{DeepHit}, a model discretizing time into intervals and jointly predicting the probability of dying in each interval, using a likelihood loss combined with a ranking loss function. \cite{gensheimer2019nnetsurvival} took a similar approach in their \emph{Nnet-survival} model (also called \emph{Logistic-Hazard} \citep{kvamme2019continuous}), predicting a conditional probability of dying within each interval, using a custom likelihood loss function. As another baseline, we consider multi-task logistic regression (\emph{MTLR}) \citep{yu2011mtlr}, which is based on a likelihood loss function optimizing a set of dependent logistic regressors for future time points. For \emph{DeepHit}, \emph{Logistic-Hazard} and \emph{MTLR}, we use an implementation using
the \verb|pycox| library. Specific details about the implementations are given in \cite{JMLR:v20:18-424}, and \cite{kvamme2019continuous}, respectively. For \emph{DeepHit} we use three layers with 240, 400, and 240 nodes, as well as parameters $\alpha=0.4$ and $\sigma=2$, and a learning rate of 3.3e-07. For
\emph{Logistic-Hazard} we use a neural network consisting of three fully-connected layers with 90, 150, and 90 nodes and a learning rate of 3.3e-07. \emph{MTLR} uses a
neural network consisting of three fully-connected layers with 90, 150, and 90 nodes and a learning rate of 3.3e-07. \emph{DeepHit}, \emph{Logistic-Hazard}, and \emph{MTLR} are discrete-time models, requiring discretization of continuous times into intervals. We use 56 intervals, each 
covering roughly 12h. All models were trained using an Adam optimizer \citep{kingma2014adam} and early stopping, terminating training when the loss on the validation set does not improve for 10 epochs.

%% file: experiments.tex
\section{Experiments}

\subsection{Data}

We use the HiRID data set \cite{hyland2019circulatoryfailure, goldberger2000hirid, faltys2020hirid}, containing clinical time series of more than 50,000 ICU admissions to a tertiary-care ICU. The data includes organ
function parameters, lab results, and treatment parameters. 
Among 197 variables, 30 important variables were chosen by analyzing their importance using mean absolute SHAP values \cite{lundberg2017unified} for the mortality prediction task,
and are listed in Table \ref{tab:used_variables}. Missing values have been imputed using forward filling and filling with a clinically normal value
if no measurement was present prior to a time-point. We use a one-hot encoding scheme to encode categorical variables and standardize all other variables to zero mean and unit variance. For further preprocessing details we refer to \cite{hyland2019circulatoryfailure}.

\begin{table}[ht!]
\caption{The subset of variables from the HiRID data set used in our experiments, in parentheses are the meta-variable IDs of the parameters as defined in \cite{hyland2019circulatoryfailure}.}
\label{tab:used_variables}
\resizebox{0.9\columnwidth}{!}{
\begin{tabular}{ lll  }
\hline
\textbf{Organ function parameters} \\
\hline
PAPm (vm9) \\ 
Cardiac output (vm13) \\ 
Heart rhythm state (vm19) \\ 
SpO$_2$ (vm20) \\ 
Respiratory rate (vm22) \\ 
Supplemental oxygen (vm23) \\ 
Urine output / time (vm24) \\ 
GCS Verbal Response / Response / Eye opening (vm25-27) \\ 
RASS (vm28) \\
Fluid output / time (vm32) \\ 
FiO$_2$ (vm58) \\
Weight (vm131) \\ 
Age \\
\hline
\textbf{Treatment parameters} \\
\hline
Norepinephrine (pm39) \\ 
Ventilator mode / Peak pressure (vm60,62) \\ 
Ventilator RR setting (vm65) \\
Propofol (pm80) \\ 
Hourly CSF drainage (vm84) \\ 
Steroids (pm91) \\ 
\hline
\textbf{Laboratory tests} \\
\hline
Arterial lactate (vm136) \\
Creatine kinase (vm144) \\ 
Mg (vm154) \\ 
Urea (vm155) \\
Bilirubine, total (vm162) \\ 
aPTT (vm166) \\ 
Total white blood cell count (vm184) \\
Platelet count (vm185) \\ 

\hline
\end{tabular}}
\end{table}

\subsection{Evaluation setup}
We drew 5 replicates of splits, each consisting of a training, a validation and a test set. Patients in the test set have a later admission time than patients in the corresponding training set, simulating model deployment on future data. We refer the reader to \cite{hyland2019circulatoryfailure} for more details on these splits. We train our models on the training set of each split. The validation set is used for early stopping, selection of optimal hyperparameters using grid search, and the analysis of variable importance. We test our models on the test set of each split, considering patients once every hour as independent test instances to estimate the future survival distribution.

\subsection{Performance comparison with baselines}
\label{sec:results}
\begin{figure}[h!]
  \centering
  \includegraphics[width=0.9\columnwidth]{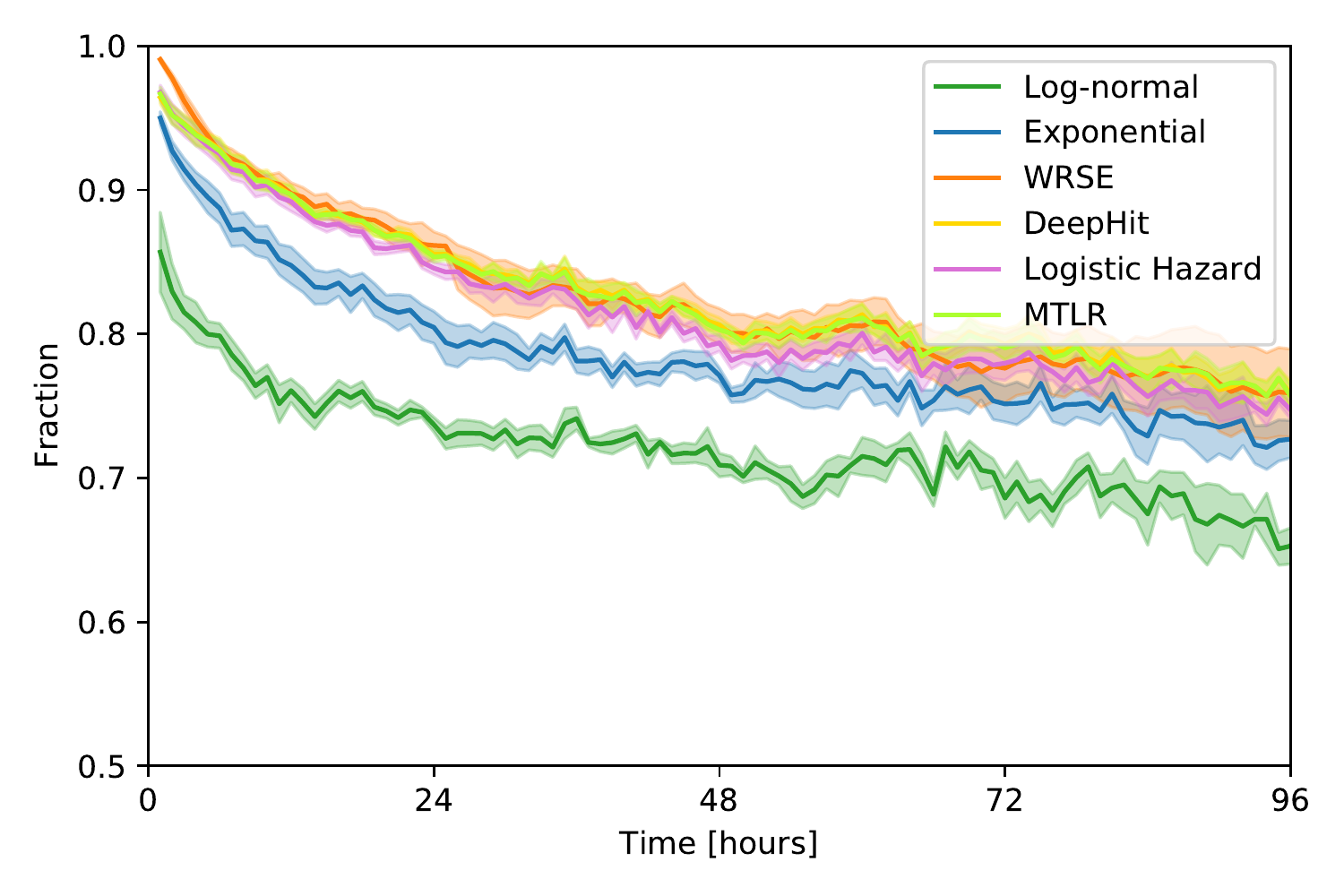}
  \caption{The fraction of concordant pairs per horizon $\tau$: $|\{(i,j) \in P_\tau | S_i(\tau) < S_j(\tau)\}| / |P_\tau|$ (cf. Section~\ref{sec:evaluation_metrics} for details on the notation). It provides an indication of discrimination performance as a function of the predictive horizon. Higher values represent better discrimination.}
  \label{fig:discrimination_time_results}
\end{figure}

\begin{figure}[h!]
  \centering
  \includegraphics[width=0.9\columnwidth]{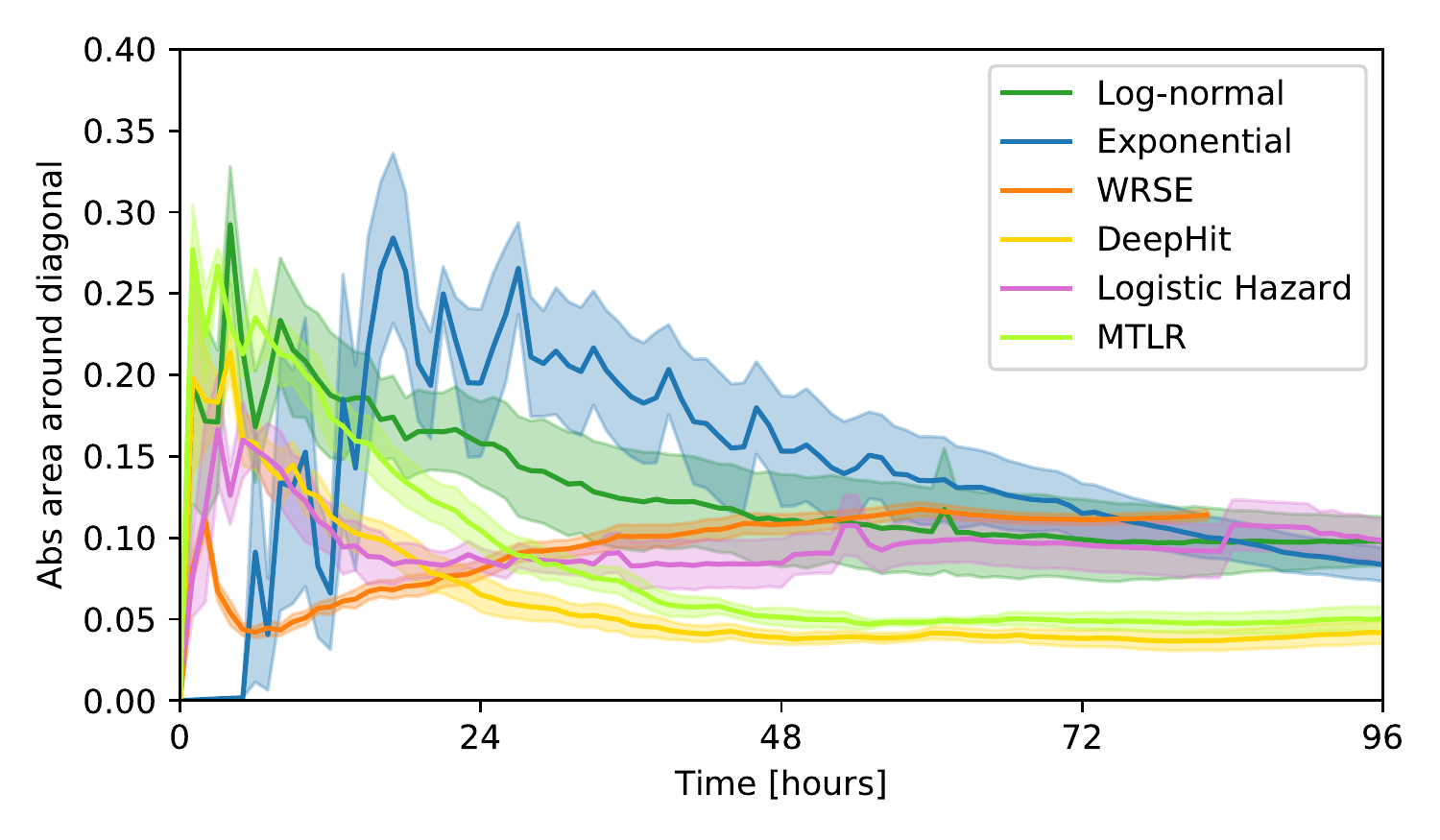}
  \caption{The area around the diagonal of the calibration plot, evaluated individually for each horizon $\tau$. Lower values correspond to better calibration.}
  \label{fig:calibration_time_results}
\end{figure}

\emph{WRSE} (with spacing given by $\gamma$=0.5) and the baseline models introduced in Section~\ref{sec:prob_time_to_death} were compared under the two weighted metrics, as described in  Section~\ref{sec:evaluation_metrics}. All models except \emph{WRSE} have been recalibrated using isotonic regression. \emph{WRSE} did not require this step, as raw predictions already show sufficient calibration. The results in Table~\ref{tab:results} demonstrate that \emph{WRSE} outperforms parametric baselines and is on par with state-of-the-art non-parametric approaches, with respect to the discriminative power ($C^{td,w}$). The trend persists across different weighting functions. Furthermore, it is better calibrated than all baselines for $\gamma$=0.3, which up-weights short-term predictions, while being outperformed by \emph{DeepHit} for $\gamma$=0.8. We complement these results by plotting the unweighted raw metrics against the time horizon $\tau$, for discrimination
in Fig.~\ref{fig:discrimination_time_results} and for calibration in Fig.~\ref{fig:calibration_time_results}. We observe that \emph{WRSE} performs competitively with the baselines for discrimination across all horizons (Fig.~\ref{fig:discrimination_time_results}).

\input{results_table}

\subsection{Training time}\label{app:computation}
The proposed \emph{WRSE} model consists of a set of independent classifiers, the training of which can be run in parallel. Isotonic regression is applied subsequently to guarantee monotonicity of the predicted 
distribution. Our result show reduced training times by a factor of 2-4x compared to the parametric baselines, and 5-9x compared to non-parametric
baselines, as depicted in Table~\ref{tab:computation}.

\begin{table}[ht!]
\caption{Average training time of the proposed model and the baselines. \emph{WRSE} runtime
refers to a parallelized ensemble, the training process runs on a multi-core CPU. 
All other models are trained on a GPU.}\label{tab:computation}.
\resizebox{0.7\columnwidth}{!}{
%\begin{tabular}{ lll  }
\begin{tabular}{ ll }
 \hline
Model & Training time [min] \\
 \hline
Log-normal & 283 \\ 
Exponential & 127 \\ 
DeepHit & 420 \\
Logistic Hazard & 349 \\
MTLR & 588 \\ 
\hline
WRSE (ours) & \textbf{63} \\ 
\hline
\end{tabular}}
\end{table}

\subsection{Analyzing different WRSE configurations}
\label{sec:comparison_of_ensemble_versions}

\begin{figure}[h!]
  \centering
  \includegraphics[width=0.9\columnwidth]{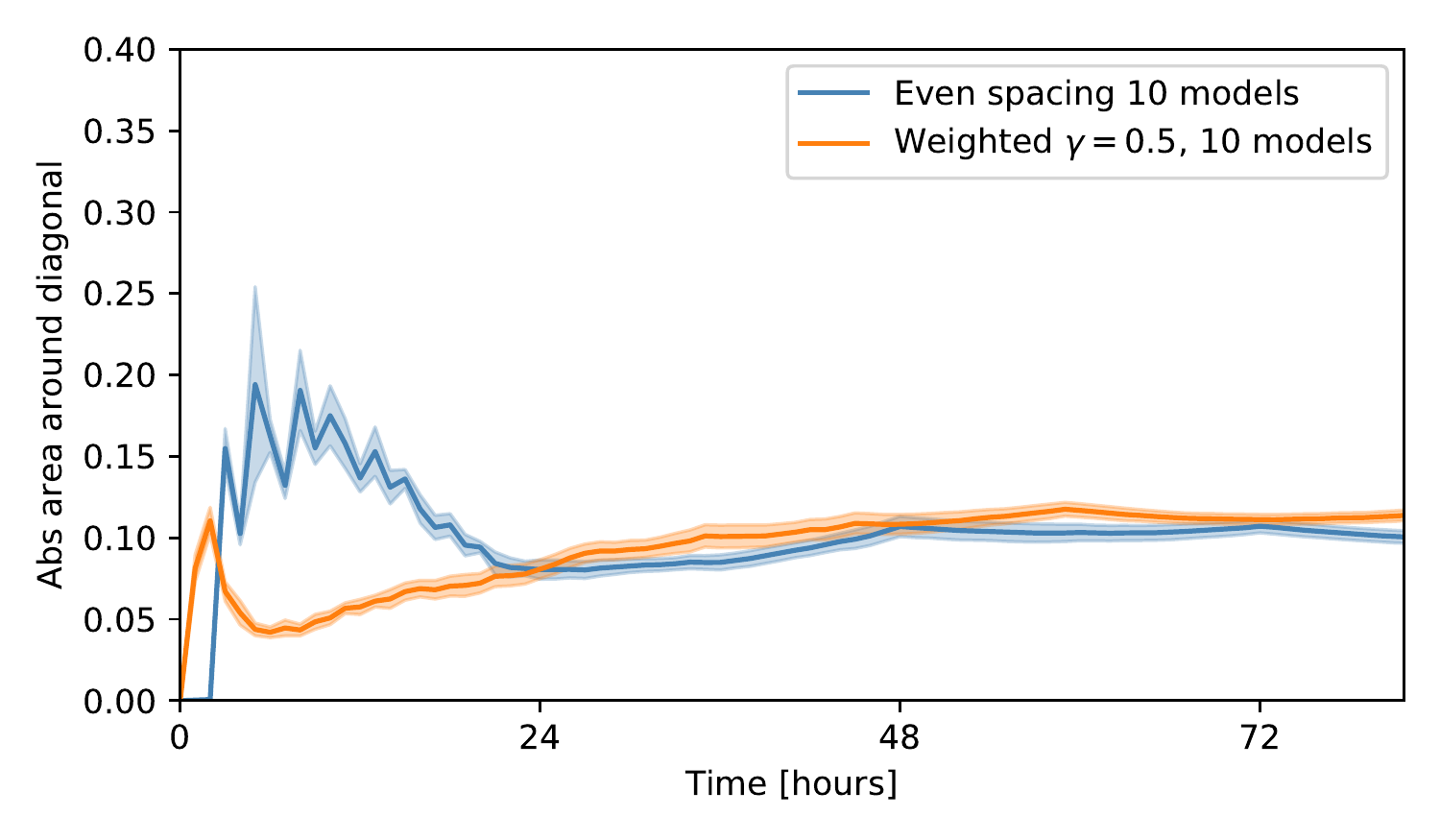}
  \caption{The area around the diagonal of the calibration plot for different variants of spacing of base models, evaluated individually for each horizon $\tau$. Lower values correspond to better calibration.}
\label{fig:calibration_ensemble_versions}
\end{figure}

To understand the effect of different weighting schemas in \emph{WRSE}, we analyze several configurations, distinguished by the number of base models (5, 7, and 10) and their spacing in time (evenly spaced vs.\ weighted spacing as described in Section~\ref{subsec:WRSE}). Results of this analysis are shown in  Table~\ref{tab:results_ensemble_version}.
We observe that the weighted versions exhibit clearly superior calibration for short-term horizons, or for long-term horizons if the number of base models is small. This is further visualized by Fig.~\ref{fig:calibration_ensemble_versions}. The discrimination performance of all models is similar, and we refer to Appendix 
Fig.~\ref{fig:concordant_pairs_ensemble_versions} for 
a closer look at the effect with respect to the horizon. In addition,
we analyze the effect of different base models in \emph{WRSE}, comparing \emph{LightGBM} against an MLP
and logistic regression. Results, given in Table \ref{tab:results_mlp_vs_lightgbm}, show that 
\emph{LightGBM} outperforms the 2 alternatives across all metrics.

\begin{table*}[htbp]
\centering
\caption{Results contrasting \emph{WRSE} with temporal weighting of base models according to $\gamma=0.5$ (cf.\ Section~\ref{subsec:WRSE}) and evenly spaced base models covering 10 days. Metrics described in Section~\ref{sec:evaluation_metrics} with three weighting functions ($\gamma \in \{0.3, 0.5, 0.8\}$) were used. More configurations are shown in Appendix Table~\ref{tab:results_ensemble_version_extended}.}

\resizebox{\textwidth}{!}{
\begin{tabular}{ lllllll  }
 \hline
 Model & $C^{td,w}$, $\gamma=0.3$ & $C^{td,w}$, $\gamma=0.5$ & $C^{td,w}$, $\gamma=0.8$ & $\text{Cal}^w$, $\gamma=0.3$ & $\text{Cal}^w$, $\gamma=0.5$ & $\text{Cal}^w$, $\gamma=0.8$ \\
 \hline
Even spacing 5 models & 0.91 $\pm$ 0.01 & 0.90 $\pm$ 0.01 & 0.87 $\pm$ 0.01 & 0.11 $\pm$ 0.01 & 0.11 $\pm$ 0.01 & 0.10 $\pm$ 0.01\\ 
Weighted $\gamma=0.5$, 5 models & 0.92 $\pm$ 0.01 & 0.90 $\pm$ 0.01 & 0.88 $\pm$ 0.01 & \textbf{0.07} $\pm$ 0.01 & \textbf{0.08} $\pm$ 0.01 & \textbf{0.08} $\pm$ 0.01\\ 
\hline 
Even spacing 7 models & 0.91 $\pm$ 0.01 & 0.90 $\pm$ 0.01 & 0.88 $\pm$ 0.01 & 0.10 $\pm$ 0.01 & 0.10 $\pm$ 0.01 & 0.09 $\pm$ 0.01\\ 
Weighted $\gamma=0.5$, 7 models & 0.92 $\pm$ 0.01 & 0.90 $\pm$ 0.01 & 0.88 $\pm$ 0.01 & \textbf{0.08} $\pm$ 0.02 & \textbf{0.08} $\pm$ 0.01 & 0.09 $\pm$ 0.01\\ 
\hline 
Even spacing 10 models & 0.91 $\pm$ 0.01 & 0.90 $\pm$ 0.01 & 0.88 $\pm$ 0.01 & 0.10 $\pm$ 0.01 & 0.10 $\pm$ 0.01 & 0.09 $\pm$ 0.01\\ 
Weighted $\gamma=0.5$, 10 models & 0.92 $\pm$ 0.01 & 0.90 $\pm$ 0.01 & 0.88 $\pm$ 0.01 & \textbf{0.07} $\pm$ 0.01 & \textbf{0.08} $\pm$ 0.01 & 0.09 $\pm$ 0.01\\ 
\hline
\end{tabular}
}
\label{tab:results_ensemble_version}
\end{table*}

\begin{table*}[htbp]
\centering
\caption{Results for \emph{WRSE} with LightGBM base models (proposed model) compared to other base models: a multi-layer perceptron (MLP) and logistic regression. More configurations are shown in Appendix Table~\ref{tab:results_mlp_vs_lightgbm_extended}. }

\resizebox{\textwidth}{!}{
\begin{tabular}{ lllllll  }
 \hline
 Model & $C^{td,w}$, $\gamma=0.3$ & $C^{td,w}$, $\gamma=0.5$ & $C^{td,w}$, $\gamma=0.8$ & $\text{Cal}^w$, $\gamma=0.3$ & $\text{Cal}^w$, $\gamma=0.5$ & $\text{Cal}^w$, $\gamma=0.8$ \\
 \hline
Weighted $\gamma=0.5$, 5 models (LightGBM) & \textbf{0.92} $\pm$ 0.01 & \textbf{0.90} $\pm$ 0.01 & \textbf{0.88} $\pm$ 0.01 & \textbf{0.07} $\pm$ 0.01 & \textbf{0.08} $\pm$ 0.01 & \textbf{0.08} $\pm$ 0.01\\ 
Weighted $\gamma=0.5$, 5 models (MLP) & 0.89 $\pm$ 0.01 & 0.88 $\pm$ 0.01 & 0.85 $\pm$ 0.01 & 0.11 $\pm$ 0.01 & 0.11 $\pm$ 0.01 & 0.10 $\pm$ 0.00\\ 
Weighted $\gamma=0.5$, 5 models (Logistic Regression) & 0.89 $\pm$ 0.01 & 0.88 $\pm$ 0.01 & 0.85 $\pm$ 0.01 & 0.11 $\pm$ 0.01 & 0.10 $\pm$ 0.01 & 0.10 $\pm$ 0.01\\ 
\hline 
Weighted $\gamma=0.5$, 7 models (LightGBM) & \textbf{0.92} $\pm$ 0.01 & \textbf{0.90} $\pm$ 0.01 & \textbf{0.88} $\pm$ 0.01 & \textbf{0.08} $\pm$ 0.02 & \textbf{0.08} $\pm$ 0.01 & 0.09 $\pm$ 0.01\\ 
Weighted $\gamma=0.5$, 7 models (MLP) & 0.89 $\pm$ 0.02 & 0.88 $\pm$ 0.02 & 0.86 $\pm$ 0.01 & 0.11 $\pm$ 0.02 & 0.10 $\pm$ 0.02 & 0.10 $\pm$ 0.02\\ 
Weighted $\gamma=0.5$, 7 models (Logistic Regression) & 0.88 $\pm$ 0.02 & 0.87 $\pm$ 0.02 & 0.85 $\pm$ 0.01 & 0.12 $\pm$ 0.04 & 0.11 $\pm$ 0.03 & 0.10 $\pm$ 0.03\\ 
\hline 
Weighted $\gamma=0.5$, 10 models (LightGBM) & \textbf{0.92} $\pm$ 0.01 & \textbf{0.90} $\pm$ 0.01 & \textbf{0.88} $\pm$ 0.01 & \textbf{0.07} $\pm$ 0.01 & \textbf{0.08} $\pm$ 0.01 & 0.09 $\pm$ 0.01\\ 
Weighted $\gamma=0.5$, 10 models (MLP) & 0.89 $\pm$ 0.01 & 0.88 $\pm$ 0.01 & 0.86 $\pm$ 0.01 & 0.11 $\pm$ 0.01 & 0.11 $\pm$ 0.01 & 0.10 $\pm$ 0.01\\ 
Weighted $\gamma=0.5$, 10 models (Logistic Regression) & 0.88 $\pm$ 0.02 & 0.87 $\pm$ 0.02 & 0.85 $\pm$ 0.01 & 0.10 $\pm$ 0.01 & 0.10 $\pm$ 0.02 & 0.10 $\pm$ 0.02\\ 
\hline
\end{tabular}
}
\label{tab:results_mlp_vs_lightgbm}
\end{table*}

\subsection{Model interpretability}
\label{subsec:model_interpretability}

By using well-established binary classification models as its basis, \emph{WRSE} remains 
interpretable to the user. It can provide global as well as per-prediction explanations of variable importance, by re-using the fast TreeSHAP algorithm \cite{lundberg2018consistent} that estimates SHAP values \cite{lundberg2017unified}. SHAP is a consistent per-prediction feature attribution approach, 
associating, for a given prediction, each feature with the conditional expectation 
of the prediction change if this feature is fixed to its actual value vs. if it was not observed. To illustrate this, we rank the variable importances of \emph{WRSE} for two different weighting functions in Table~\ref{tab:variable_importance}. Hereby, mean absolute SHAP values on the validation sets of different base models were weighted according to $\gamma$, up-weighting short-term predictions as desired. We can observe that all major organ systems, except the liver, are represented in the top variables.

\begin{table}[ht!]
\caption{Variable importance for the 20 most important variables as given by the mean absolute SHAP values \citep{lundberg2017unified} on the validation set of base models in \emph{WRSE}. We report the mean and standard deviation of the weighted mean absolute SHAP value across the five splits.}

\resizebox{\columnwidth}{!}{
\begin{tabular}{ lll  }
\hline
\textbf{Clinical parameter} & $\gamma=0.3$ & $\gamma=0.8$ \\ 
\hline
RASS (vm28) &0.31 $\pm$ 0.01  &0.30 $\pm$ 0.01 \\
Age &0.25 $\pm$ 0.02  &0.26 $\pm$ 0.02 \\
GCS Response (vm26) &0.24 $\pm$ 0.03  &0.24 $\pm$ 0.03 \\
Ventilator Peak Pressure (vm62) &0.24 $\pm$ 0.03  &0.24 $\pm$ 0.03 \\
Arterial lactate (vm136) &0.24 $\pm$ 0.02  &0.24 $\pm$ 0.02 \\
Ventilator mode (vm60) &0.18 $\pm$ 0.05  &0.18 $\pm$ 0.06 \\
GCS Verbal Response (vm25) &0.15 $\pm$ 0.03  &0.15 $\pm$ 0.03 \\
Urine output / time (vm24) &0.12 $\pm$ 0.01  &0.12 $\pm$ 0.01 \\
Weight (vm131) &0.11 $\pm$ 0.02  &0.12 $\pm$ 0.02 \\
Urea (vm155) &0.11 $\pm$ 0.02  &0.12 $\pm$ 0.02 \\
Fluid output / time (vm32) &0.11 $\pm$ 0.01  &0.11 $\pm$ 0.01 \\
SpO$_2$ (vm20) &0.11 $\pm$ 0.01  &0.10 $\pm$ 0.01 \\
Platelet count (vm185) &0.10 $\pm$ 0.00  &0.10 $\pm$ 0.00 \\
Total white blood cell count (vm184) &0.08 $\pm$ 0.00  &0.09 $\pm$ 0.00 \\
Creatine kinase (vm144) &0.08 $\pm$ 0.00  &0.08 $\pm$ 0.00 \\
Supplemental oxygen (vm23) &0.08 $\pm$ 0.02  &0.08 $\pm$ 0.02 \\
Heart rhythm state (vm19) &0.08 $\pm$ 0.01  &0.08 $\pm$ 0.01 \\
Mg (vm154) &0.07 $\pm$ 0.01  &0.07 $\pm$ 0.01 \\
GCS Eye opening (vm27) &0.07 $\pm$ 0.01  &0.07 $\pm$ 0.01 \\
aPTT (vm166) &0.06 $\pm$ 0.01  &0.07 $\pm$ 0.01 \\
FiO$_2$ (vm58) &0.07 $\pm$ 0.01  &0.06 $\pm$ 0.00 \\
\hline
\end{tabular}
}
\label{tab:variable_importance}
\end{table}

%% file: results_table.tex
\begin{table*}[htbp!]
\centering
\caption{Results of our model compared with the baselines, listing the metrics described in Section~\ref{sec:evaluation_metrics} with three weighing functions ($\gamma \in \{0.3, 0.5, 0.8\}$). We report the mean and standard error of the performance across the five splits.}

\resizebox{\textwidth}{!}{

\begin{tabular}{ lllllll  }
 \hline
 Model & $C^{td,w}$, $\gamma=0.3$ & $C^{td,w}$, $\gamma=0.5$ & $C^{td,w}$, $\gamma=0.8$ & $\text{Cal}^w$, $\gamma=0.3$ & $\text{Cal}^w$, $\gamma=0.5$ & $\text{Cal}^w$, $\gamma=0.8$ \\
 \hline
Log-normal & 0.78 $\pm$ 0.02 & 0.77 $\pm$ 0.02 & 0.75 $\pm$ 0.01 & 0.17 $\pm$ 0.05 & 0.15 $\pm$ 0.05 & 0.12 $\pm$ 0.04\\ 
Exponential & 0.87 $\pm$ 0.02 & 0.85 $\pm$ 0.02 & 0.83 $\pm$ 0.02 & 0.13 $\pm$ 0.05 & 0.14 $\pm$ 0.04 & 0.12 $\pm$ 0.03\\ 
DeepHit   & \textbf{0.91} $\pm$ 0.01 & \textbf{0.89} $\pm$ 0.01 & \textbf{0.87} $\pm$ 0.01 & 0.11 $\pm$ 0.02 & \textbf{0.09} $\pm$ 0.01 & \textbf{0.06} $\pm$ 0.01\\ 
Logistic Hazard & 0.90 $\pm$ 0.01 & \textbf{0.89} $\pm$ 0.01 & 0.86 $\pm$ 0.01 & 0.11 $\pm$ 0.02 & 0.10 $\pm$ 0.02 & 0.09 $\pm$ 0.02\\ 
MTLR           & \textbf{0.91} $\pm$ 0.01 & \textbf{0.89} $\pm$ 0.01 & \textbf{0.87} $\pm$ 0.01 & 0.16 $\pm$ 0.03 & 0.12 $\pm$ 0.02 & 0.08 $\pm$ 0.01\\ 
\hline
WRSE (ours) & \textbf{0.92} $\pm$ 0.01 & \textbf{0.90} $\pm$ 0.01 & \textbf{0.88} $\pm$ 0.01 & \textbf{0.07} $\pm$ 0.01 & \textbf{0.08} $\pm$ 0.01 & 0.09 $\pm$ 0.01\\ 
\hline
\end{tabular}

}

\label{tab:results}
\end{table*}

%% file: discussion.tex
\section{Discussion}

We proposed a non-parametric ensemble, \emph{Weighted Resolution Survival Ensemble} (\emph{WRSE}), for dynamic 
individual survival predictions in the ICU. In its default setting ($\gamma$=0.5), it allocates 
more classifiers for short-term predictions, which are more
relevant in several scenarios in an ICU. Comparisons against 
various parametric and non-parametric baselines show similar or superior performance. Our framework is adaptive to the user's choices via its 
weighting function. Once 
the temporal spacing is set, the base models are trivially 
parallelizable, resulting in a training time decrease of 2-9 times compared to the baselines. 
We also proposed calibration and discrimination metrics up-weighting short-term predictions. We believe that these metrics capture a model's usefulness in a dynamic ICU setting more closely. Since our model 
uses \emph{LightGBM} as its base model, it is easily 
inspectable using SHAP value analysis, allowing clinicians to inspect the features leading to predicted changes in mortality risk.
Future work will focus on evaluation in other cohorts, approaches for deciding the optimal temporal spacing, given a fixed budget of base models, as well as 
the introduction of other base models
for improved uncertainty estimation.

%% file: acknowledgements.tex
\section*{Acknowledgments}

This project was supported by the grant \#2017‐110 of the Strategic Focus Area "Personalized Health and Related Technologies (PHRT)" of the ETH Domain. 
JF/MH are partially supported by ETH core funding (to GR). MH is supported by the Grant No. 205321\_176005 of the Swiss National 
Science Foundation (to GR). We are grateful to Ximena Bonilla for proof-reading the manuscript.

%% file: appendix.tex
\section{Appendix}

% COMPLETED

\subsection{Additional results}

Results contrasting the different feature extractor choices
for the parametric baselines are shown in Table~\ref{tab:results_feature_extractor}. The TCN-based log-normal 
model shows better calibration and discrimination than its MLP-based counterpart. However, calibration is clearly worse. As a result, we consider the MLP-based model superior, as uncalibrated 
results are useless in practice. The exponential models are very similar; for consistency and simplicity, we define the MLP-based version as our model of choice to be included in the main results in Section~\ref{sec:results}.

\begin{figure}[h!]
  \centering
  \includegraphics[width=\columnwidth]{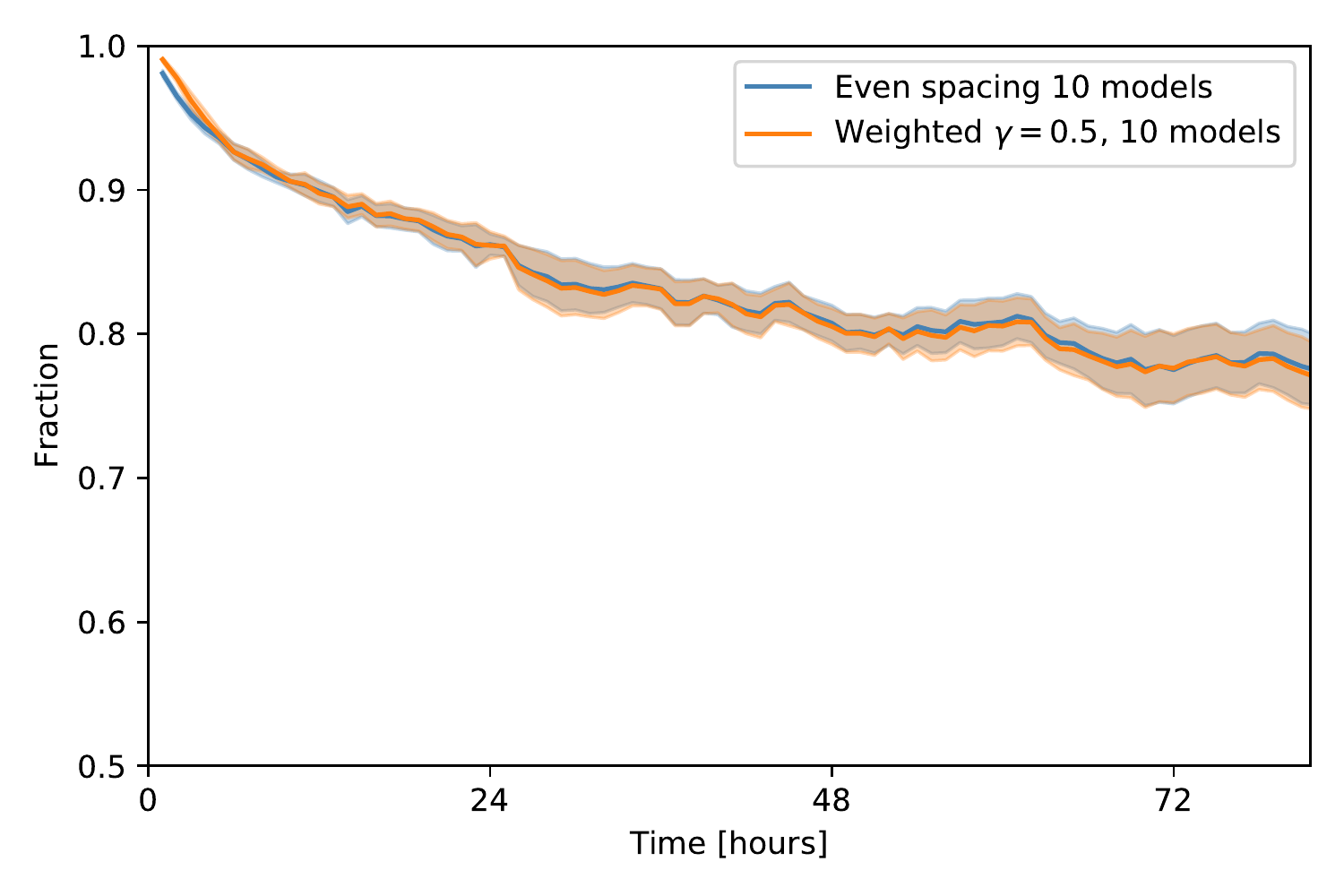}
  \caption{The fraction of concordant pairs per time $\tau$: $|\{(i,j) \in P_\tau | S_i(\tau) < S_j(\tau)\}| / |P_\tau|$ for different spacings of base models. This gives an indication of discrimination performance as a function of the predictive horizon. Higher values represent better discrimination.}
  \label{fig:concordant_pairs_ensemble_versions}
\end{figure}

An analysis of the effect of weighted temporal spacing on
the discrimination index by time horizon $\tau$ is shown
in Fig.~\ref{fig:concordant_pairs_ensemble_versions}. It is 
apparent that temporal spacing has no effect on temporal
discrimination, in contrast to calibration, as shown in the
main paper.

\begin{table*}[htbp!]
\caption{Performance of parametric models (log-normal and exponential) with two different feature extractors. The first is a multi-layer perceptron (MLP) on the current time-point. As a second feature extractor, we use a temporal convolution network (TCN) architecture on 24h of patient history.}
\resizebox{0.6\textwidth}{!}{
\begin{tabular}{ lllll  }
 \hline
 Model & $C^{td,w}$, $\gamma=0.3$ & $C^{td,w}$, $\gamma=0.8$ & $\text{Cal}^w$, $\gamma=0.3$ & $\text{Cal}^w$, $\gamma=0.8$ \\
 \hline
Log-normal MLP & 0.78 $\pm$ 0.02 & 0.75 $\pm$ 0.01 & \textbf{0.16} $\pm$ 0.05 & \textbf{0.12} $\pm$ 0.04\\ 
Log-normal TCN & \textbf{0.89} $\pm$ 0.03 & \textbf{0.83} $\pm$ 0.02 & 0.21 $\pm$ 0.01 & \textbf{0.14} $\pm$ 0.01 \\ 
Exponential MLP & \textbf{0.87} $\pm$ 0.02 & \textbf{0.83} $\pm$ 0.02 & \textbf{0.12} $\pm$ 0.04 & \textbf{0.12} $\pm$ 0.03 \\
Exponential TCN & \textbf{0.88} $\pm$ 0.02 & \textbf{0.84} $\pm$ 0.03 & \textbf{0.16} $\pm$ 0.03 & \textbf{0.11} $\pm$ 0.01 \\ 
\hline
\end{tabular}
}
\label{tab:results_feature_extractor}
\end{table*}

\begin{table*}[htbp]
\centering
\caption{Results contrasting \emph{WRSE} with different temporal spacings of base models. Metrics described in Section~\ref{sec:evaluation_metrics}  with three weighing functions ($\gamma \in \{0.3, 0.5, 0.8\}$) are shown. }

\resizebox{\textwidth}{!}{
\begin{tabular}{ lllllll  }
 \hline
 Model & $C^{td,w}$, $\gamma=0.3$ & $C^{td,w}$, $\gamma=0.5$ & $C^{td,w}$, $\gamma=0.8$ & $\text{Cal}^w$, $\gamma=0.3$ & $\text{Cal}^w$, $\gamma=0.5$ & $\text{Cal}^w$, $\gamma=0.8$ \\
 \hline
Even spacing 5 models & 0.91 $\pm$ 0.01 & 0.90 $\pm$ 0.01 & 0.87 $\pm$ 0.01 & 0.11 $\pm$ 0.01 & 0.11 $\pm$ 0.01 & 0.10 $\pm$ 0.01\\ 
Weighted $\gamma=0.3$, 5 models & 0.92 $\pm$ 0.01 & 0.90 $\pm$ 0.01 & 0.88 $\pm$ 0.01 & 0.07 $\pm$ 0.01 & 0.07 $\pm$ 0.01 & 0.07 $\pm$ 0.01\\ 
Weighted $\gamma=0.5$, 5 models & 0.92 $\pm$ 0.01 & 0.90 $\pm$ 0.01 & 0.88 $\pm$ 0.01 & 0.07 $\pm$ 0.01 & 0.08 $\pm$ 0.01 & 0.08 $\pm$ 0.01\\ 
Weighted $\gamma=0.8$, 5 models & 0.92 $\pm$ 0.01 & 0.90 $\pm$ 0.01 & 0.88 $\pm$ 0.01 & 0.09 $\pm$ 0.01 & 0.09 $\pm$ 0.01 & 0.09 $\pm$ 0.01\\ 
\hline 
Even spacing 7 models & 0.91 $\pm$ 0.01 & 0.90 $\pm$ 0.01 & 0.88 $\pm$ 0.01 & 0.10 $\pm$ 0.01 & 0.10 $\pm$ 0.01 & 0.09 $\pm$ 0.01\\ 
Weighted $\gamma=0.3$, 7 models & 0.92 $\pm$ 0.01 & 0.90 $\pm$ 0.01 & 0.88 $\pm$ 0.01 & 0.07 $\pm$ 0.01 & 0.07 $\pm$ 0.01 & 0.07 $\pm$ 0.01\\ 
Weighted $\gamma=0.5$, 7 models & 0.92 $\pm$ 0.01 & 0.90 $\pm$ 0.01 & 0.88 $\pm$ 0.01 & 0.08 $\pm$ 0.02 & 0.08 $\pm$ 0.01 & 0.09 $\pm$ 0.01\\ 
Weighted $\gamma=0.8$, 7 models & 0.92 $\pm$ 0.01 & 0.90 $\pm$ 0.01 & 0.88 $\pm$ 0.01 & 0.09 $\pm$ 0.01 & 0.09 $\pm$ 0.01 & 0.09 $\pm$ 0.01\\ 
\hline 
Even spacing 10 models & 0.91 $\pm$ 0.01 & 0.90 $\pm$ 0.01 & 0.88 $\pm$ 0.01 & 0.10 $\pm$ 0.01 & 0.10 $\pm$ 0.01 & 0.09 $\pm$ 0.01\\ 
Weighted $\gamma=0.3$, 10 models & 0.92 $\pm$ 0.01 & 0.90 $\pm$ 0.01 & 0.88 $\pm$ 0.01 & 0.07 $\pm$ 0.01 & 0.07 $\pm$ 0.01 & 0.08 $\pm$ 0.01\\ 
Weighted $\gamma=0.5$, 10 models & 0.92 $\pm$ 0.01 & 0.90 $\pm$ 0.01 & 0.88 $\pm$ 0.01 & 0.07 $\pm$ 0.01 & 0.08 $\pm$ 0.01 & 0.09 $\pm$ 0.01\\ 
Weighted $\gamma=0.8$, 10 models & 0.92 $\pm$ 0.01 & 0.90 $\pm$ 0.01 & 0.88 $\pm$ 0.01 & 0.08 $\pm$ 0.01 & 0.08 $\pm$ 0.01 & 0.09 $\pm$ 0.01\\ 
\hline
\end{tabular}

}
\label{tab:results_ensemble_version_extended}
\end{table*}

\begin{table*}[htbp]
\centering
\caption{Results for \emph{WRSE} with LightGBM (proposed model) and two alternative base models: A multi-layer perceptron (MLP) and logistic regression.}
\resizebox{\textwidth}{!}{
\begin{tabular}{ lllllll  }
 \hline
 Model & $C^{td,w}$, $\gamma=0.3$ & $C^{td,w}$, $\gamma=0.5$ & $C^{td,w}$, $\gamma=0.8$ & $\text{Cal}^w$, $\gamma=0.3$ & $\text{Cal}^w$, $\gamma=0.5$ & $\text{Cal}^w$, $\gamma=0.8$ \\
 \hline
Weighted $\gamma=0.3$, 5 models (LightGBM) & 0.92 $\pm$ 0.01 & 0.90 $\pm$ 0.01 & 0.88 $\pm$ 0.01 & 0.07 $\pm$ 0.01 & 0.07 $\pm$ 0.01 & 0.07 $\pm$ 0.01\\ 
Weighted $\gamma=0.3$, 5 models (MLP) & 0.89 $\pm$ 0.02 & 0.88 $\pm$ 0.02 & 0.85 $\pm$ 0.01 & 0.11 $\pm$ 0.01 & 0.11 $\pm$ 0.01 & 0.10 $\pm$ 0.01\\ 
Weighted $\gamma=0.3$, 5 models (Logistic Regression) & 0.88 $\pm$ 0.01 & 0.87 $\pm$ 0.01 & 0.85 $\pm$ 0.01 & 0.13 $\pm$ 0.03 & 0.13 $\pm$ 0.03 & 0.12 $\pm$ 0.03\\ 
\hline 
Weighted $\gamma=0.5$, 5 models (LightGBM) & 0.92 $\pm$ 0.01 & 0.90 $\pm$ 0.01 & 0.88 $\pm$ 0.01 & 0.07 $\pm$ 0.01 & 0.08 $\pm$ 0.01 & 0.08 $\pm$ 0.01\\ 
Weighted $\gamma=0.5$, 5 models (MLP) & 0.89 $\pm$ 0.01 & 0.88 $\pm$ 0.01 & 0.85 $\pm$ 0.01 & 0.11 $\pm$ 0.01 & 0.11 $\pm$ 0.01 & 0.10 $\pm$ 0.00\\ 
Weighted $\gamma=0.5$, 5 models (Logistic Regression) & 0.89 $\pm$ 0.01 & 0.88 $\pm$ 0.01 & 0.85 $\pm$ 0.01 & 0.11 $\pm$ 0.01 & 0.10 $\pm$ 0.01 & 0.10 $\pm$ 0.01\\ 
\hline 
Weighted $\gamma=0.8$, 5 models (LightGBM) & 0.92 $\pm$ 0.01 & 0.90 $\pm$ 0.01 & 0.88 $\pm$ 0.01 & 0.09 $\pm$ 0.01 & 0.09 $\pm$ 0.01 & 0.09 $\pm$ 0.01\\ 
Weighted $\gamma=0.8$, 5 models (MLP) & 0.90 $\pm$ 0.01 & 0.89 $\pm$ 0.01 & 0.86 $\pm$ 0.01 & 0.11 $\pm$ 0.04 & 0.11 $\pm$ 0.03 & 0.11 $\pm$ 0.02\\ 
Weighted $\gamma=0.8$, 5 models (Logistic Regression) & 0.90 $\pm$ 0.01 & 0.88 $\pm$ 0.01 & 0.86 $\pm$ 0.01 & 0.13 $\pm$ 0.04 & 0.12 $\pm$ 0.03 & 0.12 $\pm$ 0.02\\ 
\hline\hline 
Weighted $\gamma=0.3$, 7 models (LightGBM) & 0.92 $\pm$ 0.01 & 0.90 $\pm$ 0.01 & 0.88 $\pm$ 0.01 & 0.07 $\pm$ 0.01 & 0.07 $\pm$ 0.01 & 0.07 $\pm$ 0.01\\ 
Weighted $\gamma=0.3$, 7 models (MLP) & 0.88 $\pm$ 0.02 & 0.87 $\pm$ 0.01 & 0.85 $\pm$ 0.01 & 0.12 $\pm$ 0.02 & 0.12 $\pm$ 0.02 & 0.11 $\pm$ 0.02\\ 
Weighted $\gamma=0.3$, 7 models (Logistic Regression) & 0.88 $\pm$ 0.01 & 0.87 $\pm$ 0.01 & 0.85 $\pm$ 0.01 & 0.12 $\pm$ 0.02 & 0.12 $\pm$ 0.02 & 0.11 $\pm$ 0.02\\ 
\hline 
Weighted $\gamma=0.5$, 7 models (LightGBM) & 0.92 $\pm$ 0.01 & 0.90 $\pm$ 0.01 & 0.88 $\pm$ 0.01 & 0.08 $\pm$ 0.02 & 0.08 $\pm$ 0.01 & 0.09 $\pm$ 0.01\\ 
Weighted $\gamma=0.5$, 7 models (MLP) & 0.89 $\pm$ 0.02 & 0.88 $\pm$ 0.02 & 0.86 $\pm$ 0.01 & 0.11 $\pm$ 0.02 & 0.10 $\pm$ 0.02 & 0.10 $\pm$ 0.02\\ 
Weighted $\gamma=0.5$, 7 models (Logistic Regression) & 0.88 $\pm$ 0.02 & 0.87 $\pm$ 0.02 & 0.85 $\pm$ 0.01 & 0.12 $\pm$ 0.04 & 0.11 $\pm$ 0.03 & 0.10 $\pm$ 0.03\\ 
\hline 
Weighted $\gamma=0.8$, 7 models (LightGBM) & 0.92 $\pm$ 0.01 & 0.90 $\pm$ 0.01 & 0.88 $\pm$ 0.01 & 0.09 $\pm$ 0.01 & 0.09 $\pm$ 0.01 & 0.09 $\pm$ 0.01\\ 
Weighted $\gamma=0.8$, 7 models (MLP) & 0.90 $\pm$ 0.01 & 0.89 $\pm$ 0.01 & 0.86 $\pm$ 0.01 & 0.11 $\pm$ 0.03 & 0.11 $\pm$ 0.02 & 0.12 $\pm$ 0.03\\ 
Weighted $\gamma=0.8$, 7 models (Logistic Regression) & 0.90 $\pm$ 0.01 & 0.88 $\pm$ 0.01 & 0.86 $\pm$ 0.01 & 0.12 $\pm$ 0.03 & 0.11 $\pm$ 0.02 & 0.12 $\pm$ 0.02\\ 
\hline\hline 
Weighted $\gamma=0.3$, 10 models (LightGBM) & 0.92 $\pm$ 0.01 & 0.90 $\pm$ 0.01 & 0.88 $\pm$ 0.01 & 0.07 $\pm$ 0.01 & 0.07 $\pm$ 0.01 & 0.08 $\pm$ 0.01\\ 
Weighted $\gamma=0.3$, 10 models (MLP) & 0.88 $\pm$ 0.01 & 0.87 $\pm$ 0.01 & 0.85 $\pm$ 0.01 & 0.10 $\pm$ 0.01 & 0.10 $\pm$ 0.01 & 0.10 $\pm$ 0.01\\ 
Weighted $\gamma=0.3$, 10 models (Logistic Regression) & 0.88 $\pm$ 0.01 & 0.87 $\pm$ 0.01 & 0.85 $\pm$ 0.01 & 0.11 $\pm$ 0.02 & 0.11 $\pm$ 0.02 & 0.10 $\pm$ 0.01\\ 
\hline 
Weighted $\gamma=0.5$, 10 models (LightGBM) & 0.92 $\pm$ 0.01 & 0.90 $\pm$ 0.01 & 0.88 $\pm$ 0.01 & 0.07 $\pm$ 0.01 & 0.08 $\pm$ 0.01 & 0.09 $\pm$ 0.01\\ 
Weighted $\gamma=0.5$, 10 models (MLP) & 0.89 $\pm$ 0.01 & 0.88 $\pm$ 0.01 & 0.86 $\pm$ 0.01 & 0.11 $\pm$ 0.01 & 0.11 $\pm$ 0.01 & 0.10 $\pm$ 0.01\\ 
Weighted $\gamma=0.5$, 10 models (Logistic Regression) & 0.88 $\pm$ 0.02 & 0.87 $\pm$ 0.02 & 0.85 $\pm$ 0.01 & 0.10 $\pm$ 0.01 & 0.10 $\pm$ 0.02 & 0.10 $\pm$ 0.02\\ 
\hline 
Weighted $\gamma=0.8$, 10 models (LightGBM) & 0.92 $\pm$ 0.01 & 0.90 $\pm$ 0.01 & 0.88 $\pm$ 0.01 & 0.08 $\pm$ 0.01 & 0.08 $\pm$ 0.01 & 0.09 $\pm$ 0.01\\ 
Weighted $\gamma=0.8$, 10 models (MLP) & 0.90 $\pm$ 0.01 & 0.88 $\pm$ 0.01 & 0.86 $\pm$ 0.01 & 0.11 $\pm$ 0.02 & 0.11 $\pm$ 0.02 & 0.12 $\pm$ 0.02\\ 
Weighted $\gamma=0.8$, 10 models (Logistic Regression) & 0.90 $\pm$ 0.01 & 0.88 $\pm$ 0.01 & 0.86 $\pm$ 0.01 & 0.11 $\pm$ 0.01 & 0.11 $\pm$ 0.01 & 0.12 $\pm$ 0.02\\ 
\hline
\end{tabular}
}
\label{tab:results_mlp_vs_lightgbm_extended}
\end{table*}